\documentclass[11pt]{article}

\usepackage[preprint]{acl}

\usepackage{times}
\usepackage{latexsym}

\usepackage[T1]{fontenc}

\usepackage[utf8]{inputenc}

\usepackage{CJKutf8}
\newcommand{\zh}[1]{\begin{CJK}{UTF8}{gbsn}#1\end{CJK}}

\usepackage{microtype}

\usepackage{inconsolata}

\usepackage{graphicx}

\usepackage{inconsolata}

\usepackage{booktabs}
\usepackage{tabularx}
\usepackage{threeparttable}
\usepackage{amsmath}
\newcolumntype{R}{>{\raggedleft\arraybackslash}X}
\title{\textit{Pseudowords as probes}: Large Language Models show little of the sublexical sensitivity that governs human pseudoword processing}

\author{
  \textbf{Jing Chen\textsuperscript{1}},
  \textbf{Giulia Loca\textsuperscript{1}},
  \textbf{Simona Amenta\textsuperscript{2}},
  \textbf{Marco Marelli\textsuperscript{1}}
\\[6pt]
  \normalsize\textsuperscript{1}Department of Psychology, University of Milano-Bicocca, \\
  \normalsize Piazza dell'Ateneo Nuovo 1, 20126 Milano, Italy \\
  \normalsize\textsuperscript{2}Department of Informatics, Systems and Communication -- DISCo, \\
  \normalsize University of Milano-Bicocca, Milano, Italy \\[4pt]
  \normalsize\texttt{jing.chen@unimib.it}
}

\begin{document}
\maketitle
\begin{abstract}
Systematicity, the probabilistic mapping of form to meaning, permeates language at all levels, and sublexical cues have been shown to govern human pseudoword processing. Yet whether LLMs exhibit comparable sensitivity to these cues remains unclear. We tested five LLMs on two Italian two-alternative forced-choice pseudoword experiments and compared their responses with a human behavioural baseline. LLMs aligned more reliably with humans when real-word options provided a lexical familiarity cue than in the pseudoword-only condition, where they fell substantially below \textit{fastText}, a character-$n$-gram model. In addition, the sublexical cosine-similarity cue that reliably drove human--\textit{fastText} agreement did not consistently transfer to human--LLM alignment, and reasoning-token expenditure bore no consistent relation to human processing difficulty. These findings suggest that LLMs do not necessarily share the sublexical cues that govern human pseudoword processing; we discuss tokenization and training-data coverage as candidate explanations.
\end{abstract}

\section{Introduction}

Language is often characterized as an arbitrary system, where the relationship between form and meaning is treated as a matter of convention \citep{desaussure1916course, hockett1960origin}. A growing body of work shows that this arbitrariness is far from absolute (for a review, see \citealt{dingemanse2015arbitrariness}): at least two forms of non-arbitrariness are well documented, namely iconicity, i.e., the resemblance between a linguistic symbol and its referent \citep{emmorey2014iconicity, perniss2010iconicity}, and systematicity, i.e., the statistical regularities in forms that cue related meanings \citep{monaghan2014arbitrary, tamariz2008exploring, bergen2004psychological}. Systematicity in particular permeates language at all levels \citep{nolle2018emergence}, from syntactic structure (e.g., SVO) to sublexical regularities such as the phonaestheme \textit{gl-} clusters to imply things that shine (\textit{glitter}, \textit{glow}) \citep{bergen2004psychological}. 

These statistical regularities extend beyond the known lexicon to shape the processing of \textbf{pseudowords}: letter strings that conform to the orthotactic patterns of a language but carry no conventionalised meaning \citep{martinez2025pseudowords}. Although pseudowords lack any lexical entry, the sublexical form--meaning regularities of the language nonetheless guide their processing, as evidenced by \textit{fastText} \citep{bojanowski2017enriching} that reliably predict human pseudoword judgments \citep{gatti2024humans, gatti2023out, bonandrini2023form, bonandrini2026brain}.

Large language models (LLMs), as machine learning systems trained on massive corpora (typically trillions of tokens; e.g., \citealt{liu2024deepseek, achiam2023gpt}) via objectives such as next-token prediction (e.g., GPT-4) or multi-token prediction (e.g., DeepSeek-V3), also acquire rich and implicit statistical regularities from text \citep{brown2020language}. The scale and architectural sophistication of this training enables human-like performance across a wide range of linguistic and psychological tasks \citep{martinez2025ai, brysbaert2024moving, demszky2023using, srivastava2023beyond}. 

Whether the statistical regularities LLMs acquire extend to the sublexical patterns that govern human form-to-meaning mapping is, however, unclear. LLMs process text via subword tokenization (e.g., byte-pair encoding; \citealt{sennrich-etal-2016-neural}), which segments strings into tokens that are purely frequency-driven: the processing units of machine and human are thus not necessarily matched. Such investigation are especially interesting in the context of pseudowords, where they are as out-of-vocabulary (OOV) strings with no lexical entry, and thus eliminate confounds from stored word knowledge and isolate sensitivity to sublexical form cues.  

In this study, we use two Italian pseudoword experiments from \citet{gatti2024humans} as benchmarks to compare the performance of five LLMs. We ask three questions: (1)~Do LLMs align with human participants above chance level? (2)~Do sublexical features that predict humans performance also drive LLMs performance? (3)~Does computational effort in reasoning models reflect human processing difficulty? Together, we use pseudowords as probes to explore whether LLMs track the sublexical regularities that guide human pseudoword processing.

\section{Related Work}


Human pseudoword processing is sensitive to sublexical form--meaning associations already mapped in semantic memory \citep[e.g.,][]{gatti2023out}. This implicit statistical regularity is often computationally quantified by \textit{fastText} \citep{bojanowski2017enriching}, a model that represents each string as the aggregation of the vectors of its character $n$-grams, learned from distributional co-occurrence in corpora.

For example, \citet{gatti2024humans} showed that human intuitions about pseudoword meanings in two-alternative forced-choice (2AFC) tasks consistently aligned with \textit{fastText} predictions: participants preferentially chose the pseudoword whose vector was closer to the target word in \textit{fastText} space. In primed lexical decision tasks, higher \textit{fastText}-estimated relatedness between a prime word and a target pseudoword slowed responses and reduced accuracy, suggesting that the associative mechanisms governing word meaning also subserve pseudoword processing \citep{gatti2023out}. In addition, \textit{fastText}-derived semantics predicted lexical decisions for affixed pseudowords, and explicit morpheme combination further improved fit when pseudowords were accepted as words, but not when rejected \citep{bonandrini2023form}.

Unlike \textit{fastText}, which aggregates all overlapping character $n$-grams of a string, LLMs often segment character sequences into non-overlapping subword tokens selected by corpus frequency alone (e.g., byte-pair encoding; \citealt{sennrich-etal-2016-neural, mielke2021between}). A token may thus coincide with a whole word, with a sublexical unit that carries form--meaning cues for human readers, or with neither. For example, GPT-4o tokenizes \textit{glass} whole as \texttt{[glass]}, preserves the phonaestheme \textit{gl-} in \textit{glitter} \texttt{[gl, itter]}, but destroys it in \textit{glisten} \texttt{[g, listen]}. Pseudowords, by construction vanishingly rare in any training corpus, are typically split into multiple subword fragments---units that may coincide with neither \textit{fastText}'s exhaustive $n$-grams nor the cues human readers exploit. Among our stimuli of Italian pseudowords (Section~\ref{subsec:data}), for instance, GPT-4o carves \textit{mesetta} into \texttt{[mes, etta]}, handing the model a genuine diminutive suffix, but shreds \textit{silargloipe} into \texttt{[sil, arg, lo, ipe]}, a more fragmented segmentation in which no chunk isolates an Italian form--meaning cue, likely leaving the model little identifiable material from which to compose a meaning.

Tokenization, though blind to semantics, yields tokens that are not always semantically arbitrary. For example, words sharing a token are more semantically related than words sharing length-matched strings, and remain so even when they share no morpheme (e.g., the etymologically related \textit{purse} and \textit{bursary}), indicating that frequency-based tokenization captures form--meaning regularities beyond morphology \citep{haslett2025much}. LLMs, in turn, treat these tokens as functional sublexical units: they inflate the judged similarity of items that share a token even when no genuine linguistic constituent is shared (e.g., \zh{骆} `camel' and \zh{骶} `tailbone', which share their initial token but contain the semantic radicals \zh{马} `horse' and \zh{骨} `bone', respectively), and overlook shared constituents that do not surface as tokens (e.g., French \textit{entrons} is misjudged as unlike \textit{portons} and \textit{sortons} because its suffix \textit{-ons} is absorbed into the token \texttt{rons}) \citep{haslett2025tokenization}. For pseudowords, which have no lexical entry to fall back on, subword tokens are thus the primary form-based cues available to an LLM to construe meaning; \textit{whether} and \textit{how} LLMs exploit them are the questions the present study investigates.

\section{Method}
\subsection{Data and materials}
\label{subsec:data}
As a case study in Italian, we used data from the 2AFC experiments reported by \citet{gatti2024humans}. In these experiments pseudowords were generated with Wuggy \citep{keuleers2010wuggy}, a multilingual tool that produces orthographic strings conforming to the orthotactic patterns of a target language. They were hence organized in item triplets based on the semantic relatedness with attested Italian words, quantified by \textit{fastText} vectors \citep{bojanowski2017enriching}. 

In Experiment~1 (EXP1), 60 native Italian speakers (53 female, 7 male; age range: 19--34 years) responded to 50 trials. On each trial, a real target word appeared alongside a semantically related (\textit{pw\_rel}; cosine 0.20--0.44) and a semantically unrelated (\textit{pw\_unrel}; cosine $\leq 0.05$) pseudoword. The two pseudowords in each trial were matched for orthographic length and Levenshtein distance.

In Experiment~2 (EXP2), a separate group of 60 native Italian speakers (38 female, 22 male; age range: 19--35 years) completed 50 trials, in which an Italian pseudoword appeared alongside one related (cosine 0.20--0.44) and one unrelated real word (foil cosine $\leq 0.13$). The two words were matched for word and lemma frequency, orthographic length, and Levenshtein distance. Additionally, grammatical gender, number, and part of speech were also matched to prevent reliance on morphosyntactic cues.

In both experiments, participants chose the option they intuitively considered more semantically similar to the target, with reaction times recorded. Trials with reaction times below 500~ms or above 50{,}000~ms were excluded, retaining 2{,}975 trials in EXP1 and 2{,}997 trials in EXP2.

\subsection{Models}
We administered the two experiments described in Section~\ref{subsec:data} to five LLMs: Minerva-7B-instruct (henceforth Minerva; \citealt{orlando-etal-2024-minerva}), GPT-4o \citep{hurst2024gpt}, DeepSeek-V3 \citep{liu2024deepseek}, o4-mini, and DeepSeek-R1 \citep{guo2025deepseek}.

To the best of our knowledge, Minerva is the first LLM family pre-trained from scratch with Italian as a primary language, rather than adapted from an English-centric base.\footnote{\url{https://minerva-ai.org/blog}} The other four are frontier general-purpose models trained primarily on English. Of these, GPT-4o and DeepSeek-V3 are standard models that output a response directly, while o4-mini and DeepSeek-R1 are reasoning models that generate a chain of intermediate tokens before producing their final answer.

\subsection{Experimental settings}
All models were queried in Italian using a two-message format: a system prompt setting up the task and a user prompt presenting each individual trial.

The system prompt framed the task as a 2AFC study and asked the model to consider the potential meaning of the pseudoword before responding. Models were required to reply with only the selected item, with no explanation. The user prompts for Experiments~1 and 2 took the following form, respectively:

\begin{verbatim}
(EXP1) 
alcolismo

Sinistra: enfi
Destra: mavinchi

Scegli la pseudoparola più semanticamente 
simile. Rispondi con SOLO quella 
pseudoparola.
\end{verbatim}
\vspace{0.2em}
\textit{English: Choose the most semantically similar pseudoword. Respond with ONLY that pseudoword.}
\vspace{0.2em}
\begin{verbatim}
(EXP2)
troganerio

Sinistra: plotone
Destra: assessore

Scegli la parola più semanticamente simile. 
Rispondi con SOLO quella parola.
\end{verbatim}
\vspace{0.2em}
\textit{English: Choose the most semantically similar word. Respond with ONLY that word.}
\vspace{0.2em}

We tested two prompt conditions per experiment. In the baseline condition, the system prompt contained only the task description. In the role-based condition, the same description was preceded by a persona matching the demographic profile of the human sample, plus an explicit instruction to draw on native-speaker intuition (\textit{Usa la tua intuizione linguistica come se fossi un parlante nativo dell'italiano}).\footnote{English: use your linguistic intuition as if you were a native speaker of Italian.} The persona read:

\begin{verbatim}
Sei un parlante nativo italiano di 
età compresa tra 19 e 34 anni
con vista normale o corretta alla 
normalità. 
\end{verbatim}

\textit{English: You are a native Italian speaker between 19 and 34 years of age with normal or corrected-to-normal vision.}
\vspace{0.2em}

GPT-4o, DeepSeek-V3, and DeepSeek-R1 were queried via their respective chat-completion APIs at temperature 0 to minimise output variability; o4-mini does not support a temperature parameter and was left at its default. Minerva was deployed locally using the HuggingFace \texttt{transformers} library \citep{wolf-etal-2020-transformers} with greedy decoding (\texttt{do\_sample=False}), which is the local equivalent of zero temperature.

Each model completed three iterations of each experiment per condition, with the stimulus list reshuffled at the start of each iteration. For DeepSeek-R1 and o4-mini, we also recorded the number of reasoning tokens consumed per trial as a measure of computational effort (see Section \ref{analysis2}). 

\section{Analysis 1: Item-level human response consistency}

Prior work indicates substantial item-level variation in response consistency \citep{gatti2024humans, woolnough2025memorability}, though this has not been formally characterised. Before examining human--model alignment, we quantified this variation and assessed whether it reflects stable stimulus properties rather than sampling noise.

We quantified response consistency using binary entropy ($H_i$), where $p_i$ is the proportion of participants choosing the left-presented option on item $i$:

\begin{equation}
  H_i = -\bigl[p_i \log_2(p_i) + (1 - p_i)\log_2(1 - p_i)\bigr],
  \label{eq:entropy}
\end{equation}

\noindent with $H_i = 0$ indicating unanimous agreement and $H_i = 1$\,bit a perfectly split response.

\begin{figure}[htbp]
  \centering
  \includegraphics[width=\columnwidth]{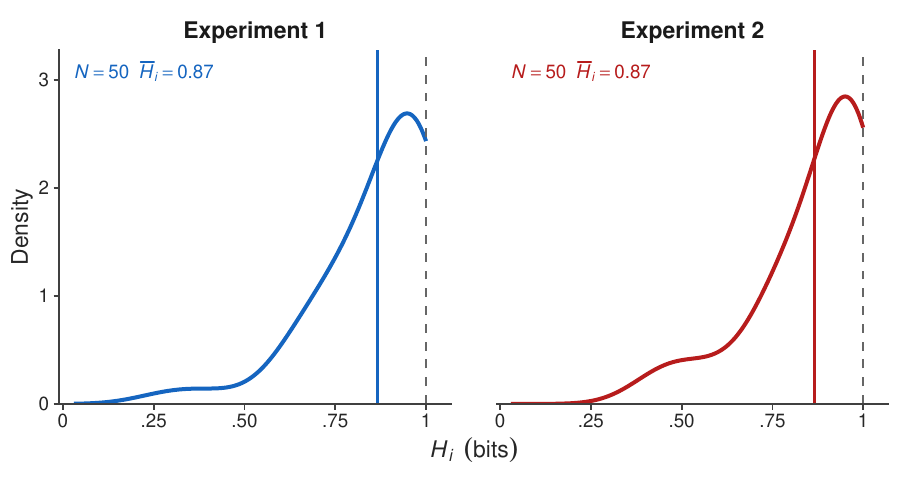}
  \caption{Distribution of item-level response entropy ($H_i$, bits) across the 50 items in each experiment. Coloured vertical lines mark the per-experiment mean ($\bar{H}_i$); the dashed grey line marks the theoretical maximum ($H = 1$\,bit).}
  \label{fig:uncertainty_distribution}
\end{figure}

As shown in Figure~\ref{fig:uncertainty_distribution}, $H_i$ values ranged from 0.290 to 1.000 in EXP1 and from 0.414 to 1.000 in EXP2, confirming substantial variation in item-level response consistency. The consistently high mean (EXP1: $M = 0.867$, $SD = 0.157$; EXP2: $M = 0.866$, $SD = 0.156$) indicates that most pseudoword pairs were difficult to reach consensus on, as expected given that pseudowords carry no conventionalised lexical meaning. 

To confirm that $H_i$ reflects a stable property of the items rather than sampling noise, we computed Spearman-Brown-corrected split-half correlations between mean item responses in two random halves of the participant sample, averaged across 1,000 splits and benchmarked against a permuted-label null. 

\begin{figure}[htbp]
  \centering
  \includegraphics[width=\columnwidth]{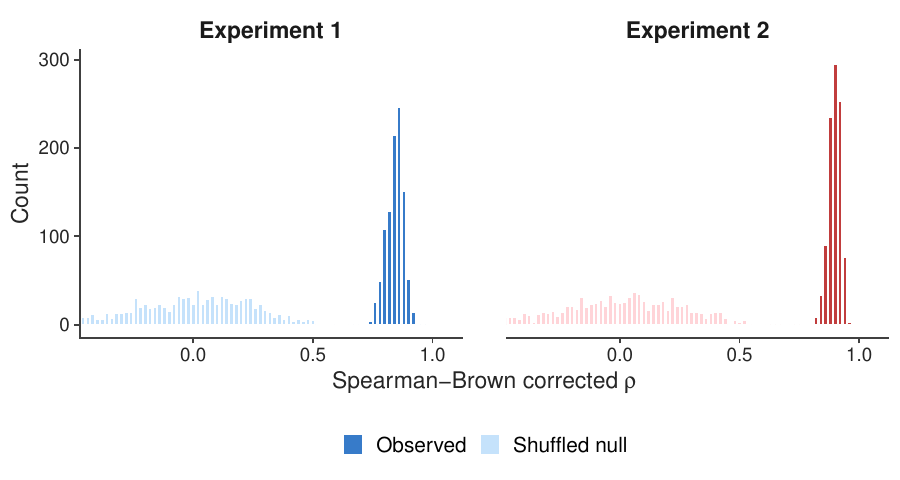}
  \caption{Split-half reliability across participants. Each panel shows the distribution of Spearman-Brown-corrected split-half correlations ($\rho$) across 1,000 random participant splits for EXP1 and EXP2. Dark bars show observed $\rho$ values; light bars show the shuffled null distribution, in which response labels were permuted before splitting.}
  \label{fig:splithalf}
\end{figure}

As shown in Figure~\ref{fig:splithalf}, the median $\rho = 0.847$ (95\% permutation interval: [0.764, 0.902]) in EXP1 and $\rho = 0.900$ ([0.840, 0.940]) in EXP2 both substantially exceeded the corresponding null medians ($\tilde{\rho}_{\mathrm{null}} = -0.005$ and $-0.006$; both $p < .001$), confirming that $H_i$ captures a genuine item-level gradient in stimulus difficulty rather than sampling noise.

Since pseudowords carry no conventionalised lexical meaning, this gradient is interpretable as variation in the strength of sublexical form cues across pairs: pairings with a stronger directional signal elicit greater consensus, whereas those with weaker or more balanced cues yield lower agreement. $H_i$ is therefore used in the subsequent analysis (Q3) as an item-level index of processing difficulty, complementing mean log response time as a measure of human effort.

\section{Analysis 2: Human-model alignment on 2AFC experiments}
\label{analysis2}

We operationalised human--model alignment as trial-level agreement between each model and individual human participants, replicating the approach used for human--\textit{fastText} alignment in \citet{gatti2024humans}. 

For each item, the three responses of any given model (one per iteration) were reduced to a single majority-vote answer. This was then compared with each participant's response on that item, yielding a binary outcome per participant--item pair: \textit{aligned} if the majority-vote answer matched the participant's choice, and \textit{misaligned} otherwise.

Minerva-7B and DeepSeek-V3 produced invalid responses on a subset of trials (e.g., non-response, multiple options), which were excluded from all analyses. For Minerva-7B, valid trial counts ranged from 2,082 to 2,577 across conditions (versus $N = 2{,}975$ in EXP1 and $N = 2{,}997$ in EXP2 for all other models). For DeepSeek-V3, exclusions were limited to the EXP2 baseline condition, retaining 2,877 of 2,997 trials.

Table~\ref{tab:model_consistency} reports within-model response consistency. Minerva-7B was the most consistent (49--58\%), likely because local greedy decoding is deterministic. Cloud API models, even at temperature~0, exhibit some run-to-run variability. 

\begin{table}[htbp]
\centering
\begin{tabularx}{\columnwidth}{Xlcc}
\toprule
Model & Prompt & EXP1 & EXP2 \\
\midrule
GPT-4o      & Baseline & 32\% & 36\% \\
            & Role     & 60\% & 44\% \\
\addlinespace[2pt]
DeepSeek-V3 & Baseline & 52\% & 40\% \\
            & Role     & 42\% & 40\% \\
\addlinespace[2pt]
DeepSeek-R1 & Baseline & 52\% & 42\% \\
            & Role     & 40\% & 32\% \\
\addlinespace[2pt]
o4-mini     & Baseline & 32\% & 20\% \\
            & Role     & 16\% & 32\% \\
\addlinespace[2pt]
Minerva-7B  & Baseline & 49\% & 58\% \\
            & Role     & 53\% & 57\% \\
\bottomrule
\end{tabularx}
\caption{Within-model response consistency: proportion of items receiving unanimous responses across three iterations.}
\label{tab:model_consistency}
\end{table}

\vspace{0.5em}
\textbf{Q1: Do LLMs align with individual participants above chance levels?}
\vspace{0.5em}

For each model, prompt condition, and the \textit{fastText} reference, we fitted an intercept-only binomial GLMM (\textit{aligned}~$\sim 1 + (1\,|\,\mathrm{ID}) + (1\,|\,\mathrm{item})$) with crossed random intercepts for participant and item. The fixed intercept estimated mean alignment against the 50\% chance level.

Table~\ref{tab:alignment} reports human-model alignment for both experiments, with the human-fastText rates (also the same from \citealt{gatti2024humans}) as reference lines. In EXP1, alignment ranged from 48.5 to 58.8\% across all models and conditions. The sole above-chance result was Minerva-7B under the baseline prompt ($58.8\%$, $p = .018$), while every other model fell short of significance. Compared to \textit{fastText} ($66.3\%$, $p < .001$), all LLMs fell substantially short.

EXP2 yielded stronger alignment overall: four models reached above-chance levels, and the best-performing model slightly exceeded the \textit{fastText} reference rate ($58.1\%$, $p = .014$): GPT-4o Role ($59.7\%$, $p = .002$), DeepSeek-V3 Base ($58.9\%$, $p = .007$), DeepSeek-R1 Role ($58.3\%$, $p = .009$), and o4-mini in both conditions ($57.8\%$, $p = .016$; $58.5\%$, $p = .008$). Minerva-7B, by contrast, fell at or below chance (44--51\%).

\begin{table*}[t]
\centering
\caption{Trial-level human-model alignment estimated from intercept-only GLMMs. $N$ = number of trials entering each fit; values are back-transformed alignment probabilities (\%). Reduced $N$ for Minerva-7B and DeepSeek-V3 (EXP2 Base) reflects exclusion of trials with invalid responses. Bold = $p < .05$.}
\label{tab:alignment}
\renewcommand{\arraystretch}{1.25}
\footnotesize
\begin{tabularx}{\textwidth}{X rrr rrr rrr rrr}
\toprule
& \multicolumn{6}{c}{\textbf{EXP 1}}
& \multicolumn{6}{c}{\textbf{EXP 2}} \\
\cmidrule(lr){2-7}\cmidrule(lr){8-13}
& \multicolumn{3}{c}{Base} & \multicolumn{3}{c}{Role}
& \multicolumn{3}{c}{Base} & \multicolumn{3}{c}{Role} \\
\cmidrule(lr){2-4}\cmidrule(lr){5-7}\cmidrule(lr){8-10}\cmidrule(lr){11-13}
Model & $N$ & \% & $p$ & $N$ & \% & $p$
      & $N$ & \% & $p$ & $N$ & \% & $p$ \\
\midrule
GPT-4o
  & 2{,}975 & 50.2 & .947
  & 2{,}975 & 53.8 & .268
  & 2{,}997 & 56.4 & .055
  & 2{,}997 & \textbf{59.7} & \textbf{.002} \\
DeepSeek-V3
  & 2{,}975 & 54.3 & .203
  & 2{,}975 & 52.2 & .520
  & 2{,}877 & \textbf{58.9} & \textbf{.007}
  & 2{,}997 & 55.5 & .099 \\
DeepSeek-R1
  & 2{,}975 & 48.5 & .665
  & 2{,}975 & 52.8 & .405
  & 2{,}997 & 55.5 & .102
  & 2{,}997 & \textbf{58.3} & \textbf{.009} \\
o4-mini
  & 2{,}975 & 56.1 & .066
  & 2{,}975 & 54.2 & .213
  & 2{,}997 & \textbf{57.8} & \textbf{.016}
  & 2{,}997 & \textbf{58.5} & \textbf{.008} \\
Minerva-7B
  & 2{,}440 & \textbf{58.8} & \textbf{.018}
  & 2{,}082 & 54.8 & .214
  & 2{,}577 & 50.7 & .853
  & 2{,}397 & 44.3 & .123 \\
\midrule
human--\textit{fastText}
  & \multicolumn{6}{c}{\textbf{66.3} ($p < .001$)}
  & \multicolumn{6}{c}{\textbf{58.1} ($p = .014$)} \\
\bottomrule
\end{tabularx}
\end{table*}

The performance reversal of Minerva-7B raised a concern about differential item coverage. Note that this model drew on fewer items in both experiments than the other models. To verify that the result was not an artefact of this difference, we rerun the GLMMs while restricting them to items with valid responses from all five models: 35 items ($N = 2{,}082$ trials) in EXP1 and 39 items ($N = 2{,}337$ trials) in EXP2. The general pattern held: Minerva-7B approached but did not reach above-chance alignment in EXP1 ($57.2\%$, $p = .052$) and fell at or below chance in EXP2 ($44.1$--$49.5\%$, both $p > .1$), while the frontier models largely replicated their EXP2 results on this restricted item set (Table~\ref{tab:alignment_shared}). 

\begin{table}[h]
\centering
\caption{Human-model alignment on the shared valid item set (items with valid responses from all five models). EXP1: $N_{\text{items}} = 35$, $N = 2{,}082$ trials; EXP2: $N_{\text{items}} = 39$, $N = 2{,}337$ trials. Values are back-transformed alignment probabilities (\%). Bold = $p < .05$.}
\label{tab:alignment_shared}
\renewcommand{\arraystretch}{1.25}
\footnotesize
\begin{tabularx}{\columnwidth}{X l rr rr}
\toprule
& & \multicolumn{2}{c}{\textbf{EXP 1}} & \multicolumn{2}{c}{\textbf{EXP 2}} \\
\cmidrule(lr){3-4}\cmidrule(lr){5-6}
Model & Prompt & \% & $p$ & \% & $p$ \\
\midrule
GPT-4o      & Base & 52.8 & .479 & \textbf{58.6} & \textbf{.022} \\
            & Role & 50.1 & .980 & \textbf{60.8} & \textbf{.003} \\
\addlinespace[2pt]
DeepSeek-V3 & Base & 50.7 & .869 & \textbf{60.1} & \textbf{.004} \\
            & Role & 47.4 & .634 & 56.1 & .116 \\
\addlinespace[2pt]
DeepSeek-R1 & Base & 46.7 & .393 & 56.4 & .091 \\
            & Role & 49.6 & .927 & \textbf{57.7} & \textbf{.037} \\
\addlinespace[2pt]
o4-mini     & Base & 54.6 & .233 & \textbf{58.6} & \textbf{.021} \\
            & Role & 51.9 & .624 & \textbf{59.6} & \textbf{.007} \\
\addlinespace[2pt]
Minerva-7B  & Base & 57.2 & .052 & 49.5 & .897 \\
            & Role & 54.8 & .215 & 44.1 & .123 \\
\bottomrule
\end{tabularx}
\end{table}

To test whether adopting the persona of a native Italian speaker improved alignment, we fitted a separate binomial GLMM for each experiment, pooling all models, with prompt condition as a fixed effect and crossed random intercepts for participant, item, and model (\textit{aligned}~$\sim \mathrm{Condition} + (1\,|\,\mathrm{ID}) + (1\,|\,\mathrm{item}) + (1\,|\,\mathrm{model})$). The role prompt had no significant effect on alignment in either experiment (EXP1: $\hat{\beta} = -0.012$, $z = -0.49$, $p = .621$; EXP2: $\hat{\beta} = -0.015$, $z = -0.61$, $p = .539$), indicating that persona framing does not systematically shift model responses toward human pseudoword judgments.


\vspace{0.5em}
\textbf{Q2: Do sublexical features predict human-model alignment here?}
\vspace{0.5em}

As shown in Q1, LLMs aligned more with humans in EXP2 than in EXP1, which is the opposite of the trend across experiments for \textit{fastText}. One potential account is that LLMs and \textit{fastText} rely on different cues.

We therefore tested whether human--LLM alignment is predicted by the same sublexical cues that \citet{gatti2024humans} entered for \textit{fastText}: $D_{\text{cos}}$ (difference in \textit{fastText} cosine similarity to the target), $D_L$ (difference in string length), and $D_{\text{LD}}$ (difference in Levenshtein edit distance to the target). In EXP2, $D_{\text{freq}}$ (difference in log word frequency) and $D_{\text{freq\_L}}$ (difference in log lemma frequency) were additionally included \citep{gatti2024humans}.\footnote{$D_{\text{cos}}$ is derived from \textit{fastText} vectors for all models, serving as item-level index of the sublexical semantic signal available in the stimuli rather than as a model-specific similarity measure.}

LLM--human alignment was modelled with binomial GLMMs via \texttt{lme4} \citep{bates2015fitting}.\footnote{Participant random intercepts were omitted; fits were singular when included.} Each predictor is an item-level difference between the two response options, computed as the value for the related option minus that for the unrelated option, so that positive values indicate an advantage for the related option on that dimension (e.g., $D_{\text{cos}} > 0$ when the related option is closer to the target in \textit{fastText} space). All predictors were $z$-scored within each experiment prior to entry. (EXP1: \textit{aligned}~$\sim D_{\text{cos}} + D_L + D_{\text{LD}} + (1\,|\,\mathrm{item})$; EXP2: \textit{aligned}~$\sim D_{\text{cos}} + D_L + D_{\text{LD}} + D_{\text{freq}} + D_{\text{freq\_L}} + (1\,|\,\mathrm{item})$).

As shown in Table~\ref{tab:predictors}, no sublexical predictor consistently predicted human--LLM alignment. In EXP1, no model reached significance on any predictor. In EXP2, only DeepSeek-R1 showed a reliable effect, for $D_{\text{cos}}$ ($\hat{\beta} = 0.35$, $p = .008$); all other models remained non-significant. The human--\textit{fastText} reference, by contrast, showed a reliable $D_{\text{cos}}$ effect in both experiments (EXP1: $\hat{\beta} = 0.20$, $p = .041$; EXP2: $\hat{\beta} = 0.27$, $p = .030$) \citep{gatti2024humans}. This effect is expected by construction, as \textit{fastText} chooses whichever option is closer in cosine terms; precisely for this reason, it serves as a positive control, confirming that the items carry a sublexical signal strong enough to be detected in this paradigm and tracked by human participants. The null effects for the LLMs therefore point to insensitivity to this cue.

\begin{table*}[t]
\centering
\caption{Item-level sublexical predictors of human--model alignment. LLM estimates pool across prompt conditions. Bold = $p < .05$; all predictors $z$-scored.}
\label{tab:predictors}
\renewcommand{\arraystretch}{1.25}
\setlength{\tabcolsep}{2pt}
\footnotesize
\begin{tabularx}{\textwidth}{X rr rr rr rr rr rr}
\toprule
& \multicolumn{2}{c}{fastText} & \multicolumn{2}{c}{DeepSeek-R1} & \multicolumn{2}{c}{DeepSeek-V3} & \multicolumn{2}{c}{GPT-4o} & \multicolumn{2}{c}{Minerva-7B} & \multicolumn{2}{c}{o4-mini} \\
\cmidrule(lr){2-3}\cmidrule(lr){4-5}\cmidrule(lr){6-7}\cmidrule(lr){8-9}\cmidrule(lr){10-11}\cmidrule(lr){12-13}
& $\hat{\beta}$ & $p$ & $\hat{\beta}$ & $p$ & $\hat{\beta}$ & $p$ & $\hat{\beta}$ & $p$ & $\hat{\beta}$ & $p$ & $\hat{\beta}$ & $p$ \\
\midrule
\multicolumn{13}{l}{\textit{EXP 1}} \\[2pt]
\quad$D_{\text{cos}}$     & $\phantom{-}\textbf{0.20}$ & $\textbf{.041}$ & $-0.08$ & $.602$ & $\phantom{-}0.12$ & $.480$ & $-0.17$ & $.282$ & $\phantom{-}0.14$ & $.395$ & $-0.13$ & $.355$ \\
\quad$D_L$                & $\phantom{-}0.31$           & $.057$          & $-0.01$ & $.952$ & $-0.01$           & $.983$ & $\phantom{-}0.11$ & $.656$ & $-0.10$           & $.708$ & $\phantom{-}0.34$ & $.146$ \\
\quad$D_{\text{LD}}$      & $-0.32$                     & $.054$          & $\phantom{-}0.12$ & $.594$ & $-0.15$ & $.624$ & $-0.12$ & $.661$ & $\phantom{-}0.13$ & $.661$ & $-0.26$ & $.285$ \\
\addlinespace
\multicolumn{13}{l}{\textit{EXP 2}} \\[2pt]
\quad$D_{\text{cos}}$     & $\phantom{-}\textbf{0.27}$  & $\textbf{.030}$ & $\phantom{-}\textbf{0.35}$ & $\textbf{.008}$ & $\phantom{-}0.19$ & $.146$ & $\phantom{-}0.12$ & $.389$ & $-0.04$ & $.787$ & $\phantom{-}0.22$ & $.132$ \\
\quad$D_L$                & $\phantom{-}0.20$            & $.137$          & $\phantom{-}0.08$ & $.552$ & $\phantom{-}0.02$ & $.860$ & $\phantom{-}0.10$ & $.537$ & $\phantom{-}0.08$ & $.637$ & $\phantom{-}0.01$ & $.968$ \\
\quad$D_{\text{LD}}$      & $-0.20$                      & $.147$          & $\phantom{-}0.03$ & $.845$ & $-0.23$           & $.109$ & $-0.11$           & $.526$ & $\phantom{-}0.23$ & $.187$ & $\phantom{-}0.11$ & $.480$ \\
\quad$D_{\text{freq}}$    & $\phantom{-}0.20$            & $.637$          & $\phantom{-}0.04$ & $.919$ & $-0.01$           & $.985$ & $-0.03$           & $.953$ & $-0.33$           & $.537$ & $\phantom{-}0.15$ & $.779$ \\
\quad$D_{\text{freq\_L}}$ & $-0.33$                      & $.453$          & $\phantom{-}0.10$ & $.823$ & $-0.15$           & $.745$ & $\phantom{-}0.10$ & $.849$ & $\phantom{-}0.51$ & $.387$ & $-0.13$           & $.798$ \\
\bottomrule
\end{tabularx}
\end{table*}

\vspace{0.5em}
\textbf{Q3: Does reasoning-model computational effort predict human processing difficulty?}
\vspace{0.5em}

We next asked whether items that demand greater cognitive effort from humans also drive greater computational effort in reasoning models.

We operationalized model effort as the reasoning tokens consumed per trial. Specifically, we took the mean token count across the three iterations as a more stable indicator of item-level processing cost for each reasoning model. Meanwhile, human processing effort was indexed by mean response time (RT) per item, with response entropy $H_i$ included as a complementary measure. 

For each reasoning model and prompt condition, we correlated log mean reasoning tokens with mean log human RT per item ($N = 50$), then recomputed these as partial correlations after residualising both variables on $D_{\text{cos}}$ to control for stimulus discriminability.

As shown in Table~\ref{tab:rt_tokens}, only one association was reliable: o4-mini (Baseline) in EXP1 ($r = .28$, $p = .046$; partial $r = .29$, $p = .039$). The direction was positive: items that took humans longer also elicited more reasoning tokens from o4-mini. This association persisted after controlling for $D_{\text{cos}}$, indicating it is not simply an artefact of stimulus discriminability. However, no other model, condition, or experiment yielded a consistent effect.

\begin{table}[t]
\centering\small
\setlength{\tabcolsep}{5pt}
\caption{Pearson correlations between LLM reasoning tokens and
         mean item-level human response time.}
\label{tab:rt_tokens}
\begin{threeparttable}
\begin{tabularx}{\columnwidth}{@{}Xcc@{}}
\toprule
Model & $r\ (p)$ & $r_{\text{part}}\ (p)$ \\
\midrule
\multicolumn{3}{@{}l}{\textit{EXP 1}}\\[2pt]
\quad DeepSeek-R1\ \ base & $-.10$\ (.494) & $-.12$\ (.424) \\
\quad DeepSeek-R1\ \ role & $\phantom{-}.21$\ (.147) & $\phantom{-}.20$\ (.170) \\
\quad o4-mini\ \ \ \ \ base   & $\phantom{-}\textbf{.28}$\ (\textbf{.046}) & $\phantom{-}\textbf{.29}$\ (\textbf{.039}) \\
\quad o4-mini\ \ \ \ \ role   & $\phantom{-}.23$\ (.116) & $\phantom{-}.26$\ (.065) \\
\addlinespace
\multicolumn{3}{@{}l}{\textit{EXP 2}}\\[2pt]
\quad DeepSeek-R1\ \ base & $\phantom{-}.02$\ (.880) & $\phantom{-}.01$\ (.968) \\
\quad DeepSeek-R1\ \ role & $-.05$\ (.730) & $-.08$\ (.589) \\
\quad o4-mini\ \ \ \ \ base   & $\phantom{-}.01$\ (.942) & $-.07$\ (.608) \\
\quad o4-mini\ \ \ \ \ role   & $\phantom{-}.16$\ (.259) & $\phantom{-}.06$\ (.679) \\
\bottomrule
\end{tabularx}
\begin{tablenotes}[flushleft]\footnotesize
\item $r_{\text{part}}$: partial correlation controlling for $\Delta_{\cos}$
      (stimulus discriminability); $N = 50$ items per cell.
\end{tablenotes}
\end{threeparttable}
\end{table}

\begin{figure}[htbp]
  \centering
  \includegraphics[width=\columnwidth]{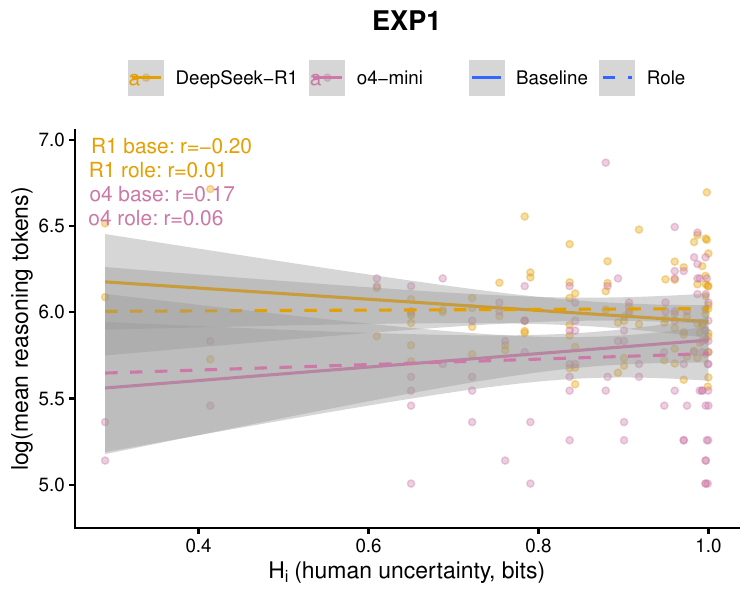}\\[4pt]
  \includegraphics[width=\columnwidth]{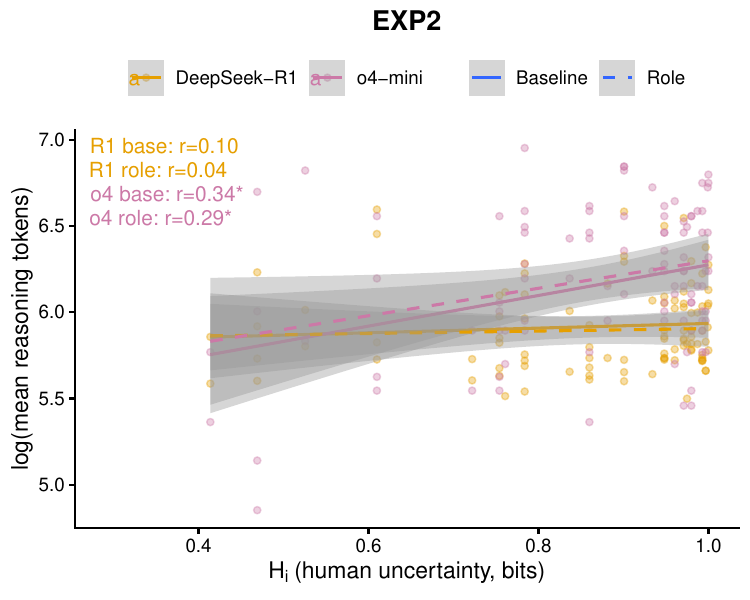}
  \caption{Association between model reasoning tokens and human response entropy in EXP1 (top) and EXP2 (bottom). Each point represents one of the 50 items, with the x-axis showing log mean reasoning tokens and the y-axis showing human response entropy ($H_i$). Lines show best-fitting linear regression for each model and prompt condition.}
  \label{fig:tokens_entropy}
\end{figure}

We also examined whether log mean reasoning tokens correlated with item-level response entropy $H_i$. As shown in Figure~\ref{fig:tokens_entropy}, only o4-mini in EXP2 produced reliable associations under both prompt conditions (Baseline: $r = .34$, $p = .015$; Role: $r = .29$, $p = .043$), with more tokens generated for items on which human responses were more evenly split. DeepSeek-R1 showed no significant correlation in either experiment, with a relatively flat slope in EXP2.

Across both effort measures, the direction of o4-mini's associations was positive: items demanding more human processing also elicited more reasoning tokens. The consistent positive direction across both measures is noteworthy, but the associations were not consistent across experiments and were absent entirely for DeepSeek-R1. The results do not, therefore, support a general correspondence between reasoning-model token expenditure and human item-level processing demands on pseudoword tasks.

\section{Conclusion}

In this study, we used pseudowords as probes to examine whether LLMs share the sublexical sensitivity that affects humans' semantic intuitions. The results suggest that, by and large, they do not.

First, LLMs did not consistently match human pseudoword-related judgments at above-chance rates, and the pattern varied across experiments and models. Specifically, alignment was generally better in EXP2, where two real-word options provided a stronger lexical familiarity cue, than in EXP1. In EXP1, where both options were pseudowords, LLMs aligned with humans at or near chance, with performance well below the \textit{fastText} reference.

Second, $D_{\text{cos}}$, the sublexical predictor that reliably drove human--\textit{fastText} agreement, showed no consistent effect on human--LLM alignment, with DeepSeek-R1 in EXP2 as the sole exception.

Third, reasoning-token counts showed no consistent relation with human processing difficulty across models: only o4-mini showed scattered positive associations, and DeepSeek-R1 showed none.

We close with two speculative and mutually compatible explanations for this insensitivity. The first is tokenization: \textit{fastText} represents a pseudoword through all of its overlapping $n$-grams, so every informative substring contributes to its meaning estimate, whereas an LLM receives a single frequency-driven segmentation, in which a form--meaning cue survives mainly when the tokenizer happens to isolate it (e.g., GPT-4o's tokenizer yields \textit{mes+etta} vs.\ \textit{sil+arg+lo+ipe}). If LLMs treat tokens as their functional sublexical units \citep{haslett2025tokenization}, regularities that do not surface as tokens remain largely invisible to them---which would explain why the $n$-gram-based $D_{\text{cos}}$ predicted human--\textit{fastText} but not human--LLM alignment. 

The second explanation is training data: the sublexical regularities of Italian may be underrepresented in predominantly English corpora. Suggestively, Minerva-7B, which is the only Italian-primary model tested, was also the only model to exceed chance in the pseudoword-only EXP1 (under the baseline prompt, and only marginally so on the shared-item set), while trailing the frontier models in EXP2, where lexical knowledge sufficed. The two accounts may further converge: token inventories are themselves induced from the same English-skewed corpora, leaving lower-resource languages with less informative segmentations \citep{haslett2025much}.

\section*{Limitations}

This study has at least two limitations. First, as discussed above, the training-data account cannot be ruled out: the observed alignment gap may reflect differences in training-data coverage as much as differences in sublexical processing, and replication in higher-resource languages would help disentangle the two.

Second, both experiments drew on 50 pseudoword pairs from the original dataset. Although the alignment analyses operated at the trial level and benefited from over 2,000 observations per fit, analyses requiring item-level aggregation (including the sublexical predictor models and the reasoning-token correlations) were constrained to 50 data points, and estimates from these analyses should therefore be interpreted with caution.

\section*{Data Availability}
All code and data are available on OSF at \href{https://osf.io/5swkp/overview?view_only=6af88d394dec4e3181891d612925d1c5}{osf.io/5swkp}.

\section*{Acknowledgments}
This project was supported by the European Union (ERC-COG-2022, BraveNewWord, 101087053). Views and opinions expressed are, however, those of the authors only and do not necessarily reflect those of the European Union or the European Research Council Executive Agency. Neither the European Union nor the granting authority can be held responsible for them.

\bibliography{custom}




\end{document}